# Using Graph Convolutional Networks and TD(λ) to play the game of Risk

by

## Jamie Carr

August 2020

# Abstract

Risk is 6 player game with significant randomness and a large game-tree complexity which poses a challenge to creating an agent to play the game effectively. Previous AIs focus on creating high-level handcrafted features determine agent decision making. In this project, I create D.A.D, A Risk agent using temporal difference reinforcement learning to train a Deep Neural Network including a Graph Convolutional Network to evaluate player positions. This is used in a game-tree to select optimal moves. This allows minimal handcrafting of knowledge into the AI, assuring input features are as low-level as possible to allow the network to extract useful and sophisticated features itself, even with the network starting from a random initialisation. I also tackle the issue of non-determinism in Risk by introducing a new method of interpreting attack moves necessary for the search. The result is an AI which wins 35% of the time versus 5 of best inbuilt AIs in Lux Delux, a Risk variant.

# Contents

# 1 Introduction

## 1.1 What is Risk

Risk is a well-known board game released in 1959. The game has different implementations and variations since on both digital and board forms. It is played with 2-6 players. With the classic map, the world is divided into 42 territories and 6 continents. The goal of the game is eliminating other players and conquer the world.

At the start of the game, there is an initial placement of armies. For this project, territories and armies are randomly distributed.

The game is split into 4 sequential stages.

1. Cards Stage: Players take a card from the deck if they conquer a territory during their turn at the end of the Fortification Stage. The card can either be an Infantry, Calvary, Artillery plus a Wildcard. If they have a matching set of 3 or one of a kind the player can cash them in for armies to use in the placing stage. If a player has 5 or more cards at any stage in the game, they must cash them in.
2. Placing Stage: Players gain income from territories and continents held. 1 extra army per 3 territories, plus a bonus from any continent(s) held, plus armies from cashed cards.
3. Attack Stage: Players then use armies to attack territories not owned by them. Attacks are determined by dice throws. Details shown in 3.3. This is the most significant factor of nondeterminism in the game, the other being cards. Each territory must have a minimum of 1 army. After a successful attack, the attacker must leave at least 3 units in the newly conquered territory, but then can decide how to distribute their armies. There is no limit on the number of attacks a player can execute. If a player kills another player, the attacker takes his cards. If the attacker has more than 5 cards, then the attacker goes to the Cards Stage. After a successful attack, the attacker chooses how many units to place in the newly conquered territory from his remaining armies with a minimum of 3.
4. Fortification Stage: If the player wants to, they can move armies from one of their territories to one of their adjacent territories, but only once, while maintaining a minimum of 1 army per territory.

The 4 stages were described with the Hasbro official ruleset in 1980. However, there are many variations which change the rules. See 3.2 for the setup I use.

## 1.2 Motivations

Compared to games like Chess and Go, Risk has added complexities in 4 main areas

- The non-determinism mainly from dice rolls means it is difficult to see more difficult to search future game states and requires constant re-evaluation to resync estimates with observed behaviour. It is difficult for an agent to predict into the future.
- There are more than 2 players.
- The number of moves an agent can execute in one turn is not capped
- Partially due to the above, Risk has a very large state space and game tree complexity, see 2.2

Previous attempts at a Risk AI, see 2.2, rely on many handcrafted heuristics with limited parameter tuning through learning to handle these complexities. For D.A.D, I aim to design an architecture and system to allow the AI to learn through observing games at a low level with minimal feature engineering to reduce handcrafted knowledge built into the AI. I was inspired by the motivations behind the Giraffe chess engine [1] and the success of AlphaGo in Chess and Go.[2]

# 2 Related Work

## 2.1 Lux Delux

Lux Delux is a paid Risk variant written in Java [3]. Users can create custom plugins and maps including AIs. In Lux, fortifications can occur an unlimited amount of times, but only to units previously not fortified during that turn. There are many different card progression sequences to choose from, including one static sequence where cashing cards will always yield 5 extra units.

I will be using Lux to implement my AI because:

- Has a superfast mode. This is useful to collecting AI game data for training.
- Has a variety of difficult AIs to compare D.A.D against.
- Comes with an SDK where I can subclass the inbuilt AIs to collect game state data and statistics for evaluation
- Configurable to different Risk variants.

## 2.2 Other Risk AIs

F. Olsson showed that a multi-agent system utilising distributed problem solving can play Risk with progressive cards successfully, evaluating above the inbuilt Lux AIs. However, it relies entirely on high-level handcrafted features.[4]

M. Lütolf created an AI with Temporal Difference learning and handcrafted features however the learning failed to increase the skill of the agent from a human initialisation. Lütolf also analysed the Lux AIs and the methodology behind them, showing how AIs were created from a chain of principles, with harder AIs being children of simpler, easier AIs. [5]

M. Wolf's AI implemented human defined high-level goals with handcrafted features and was evaluated versus human players with success. Temporal Difference was applied with moderate success at improving the skill of the agent. Wolf also performed game-tree and state-space analysis, showing Risk's game-tree complexity to far exceed that of even Chess or Go. This is also evidence for his claim that "even searching a game tree of decisions of a single game turn is infeasible". I show that it is possible to construct partial game-trees with search heuristics that are deep enough to execute complex attack sequences in 3.5 and 3.8 [6]

Lux Delux comes with a variety of inbuilt AIs, most of which are described by F. Olsson [4]. The overall concept is to build AIs from a set of human-defined principles, then extend their functionality through inheritance to add and refine those principles.

These AIs have used hand-crafted features to guide the agent's decision making. None of the AIs used a search in any capacity with Wolf citing the game-tree complexity. This means none of the AIs can see in the future. This is overcome using high-level handcrafted features which engineer human knowledge into the AI. The goal of my project is to see if an agent can learn with low-level features and minimal human knowledge. Solving the issue with constructing a game-tree and creating an evaluation function sophisticated enough to extract high level features are main goals in this project.

# 3 D.A.D: A Risk Playing Agent

## 3.1 Overview

The AI operates on the following algorithm:

1. If it is the AIs turn to play, read the game state from Lux.
2. With a game tree search, attempt to explore all turn end-states and return the move chain to reach the predicted game state which has the highest score measured by the evaluation function.
3. If there are no attack moves, send them to the game server. Else send all moves up to the first attack inclusive. Go to step 1.

It is important to not send moves after an attack because the attack is non-deterministic. The AI will re-sync with the game server after the attack. This can be inefficient because similar game-states are explored multiple times. This could be fixed by checking to see if the attack result fell within an expected range then sending the previously searched moves after the attack. However, I did not implement this in the project because it would make debugging harder and the focus of this project is not to create a fully time-optimised AI, rather to show that an AI built on the concepts described in the 1.2 and 2.1 can play at a high level.

## 3.2 Setup

The AI will initially be created to play Risk with a static 5-5-5 card sequence and non-increasing continent bonuses. Australia, South America, Africa, America, Europe, Asia having bonuses of 2, 2, 3 5, 5 and 7 respectively. All games will feature 6 players. Fortifications are done as in Lux Delux.

## 3.3 Attack Handler

When a player attacks in Risk, they throw dice to determine the next stage of the attack. I refer to this as a *dice outcome*. The attacker gets 1 dice per attacking unit, with a maximum of 3. The defender gets 1 dice per defending unit, with a maximum of 2. The attackers highest dice roll is compared to the defender's highest dice roll, then next highest with the defender's other dice. For every dice, If the attacker's dice is higher than the defender's dice, then the defender loses 1 army, else the attacker loses 1 army.

In this project, I define an *attack* to be when all attacking units from one territory engage the defending units of another. I define the *terminal state* of an attack to be where either the defending or attacking armies are equal to zero. Evaluating an attack means to consider an attack in terms of its terminal state.

One of the main issues for creating an AI in Risk is the non-determinism caused by rolling dice for attacks. This means it is hard to construct a game-tree because it becomes extremely large with all possibilities. I create the Attack Handler to consider attacks in risk deterministically by considering a specific terminal state. This is inspired by when people play Risk. They typically consider a certain outcome for their attack sequence learned from experience, then update this knowledge after every attack. I implement a similar approach with the AI, modelling the game deterministically, then resyncing the model after every attack is performed on the game server.

To evaluate a given attack to a terminal state, I first pre-calculate all possible terminal states, then select a specific one. I use the following algorithm to populate a table for all possible terminal states for all possible attacks (up to a cap):

1. Let $A, D$ be the starting number of attackers and defenders in an attack. Let $R$ represent the ordered list of terminal states from best for the defender to best for the attacker, initially empty which will be added to the table.
2. Create unweighted tree-like graph, $G$, with each node representing a possible future state of the attack containing:
    1. Number of attackers, $a$
    2. Number of defenders, $d$
    3. Probability that this state would occur, $p$
3. Initialise the root node with $a = A, d = D, p = 1$.
4. For each leaf node, $l$, create a child, $c$, for every dice outcome possible from the parent position.
$$a_c = a_l - a_{lost}, d_c = d_l - d_{lost}, p_c = p_l * p_{outcome}$$
5. If no child nodes were created, stop.
6. For all the newly created children, merge m child nodes that have the same $a$ and $d$. For each merged child:

$$p_c = \sum_{i=0}^{i=m} p_{c_i}$$

7. For every child, if it has reached a terminal state, add to $R$ and remove from $G$. Go to step 3.

Once the lookup table is completed, for a given attack, we can find $R$, which is the list of size $s$ containing every terminal state, $t$. The AI considers the terminal state with the highest value of $t$ which satisfies:

$$Risky \leq \sum_{t=0}^{s} p_t$$

Where Risky is a parameter between 0 and 1 used to control how optimistic the AI is when evaluating attacks.

It is possible that a very large attack will occur where there is no list of terminal states precalculated. In this case, it is calculated dynamically.

## 3.4 Neural Network Evaluation function
### 3.4.1 Network architecture

An evaluation function is required to evaluate turn end-states from a game-tree to select optimal moves with a tree search, using a subset of the legal moves defined by heuristics, see 3.5

Because the board can be represented as an unweighted, undirected graph, a Graph Convolutional Network (GCN) [7] can operate directly on the board and extract local features which can generalise across the board. The GCN is analogous to a Convolutional Neural Network where they operate on graphs as opposed to tensor data. However, it is not efficient to represent global features such as total armies, income and territories of each player. A fully connected layer (FC1) is used to process global features. I use FC 2 to mix the signals from GCN 2 and incorporate node specific knowledge. I then concatenate FC2 and FC1, processing them with FC3 to "mix" Global and Board knowledge together before adding the output layer.

This architecture allows the network to extract sophisticated features from low-level inputs automatically and hence does not require handcrafting features and integrating human knowledge into the AI. This means the AI acquires much of the knowledge about the game through learning itself.

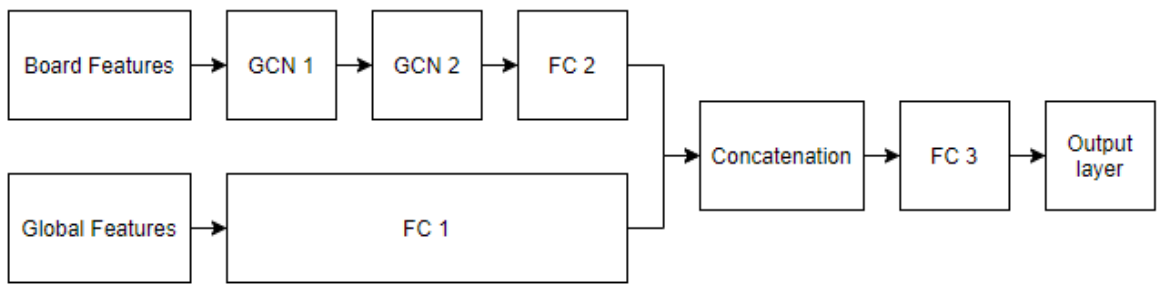

Figure 1: Network architecture

After each layer I apply a ReLU activation function except for the output layer where there is none. Using a Softmax on the output layer to predict each players probability of winning the game sounds desirable however I observed the AI to struggle in positions where it was essentially won because the difference in the evaluation between its current position and a won position was insignificant.

The network is trained to evaluate all 6 player positions on the board. This is efficient because information within hidden layers can be shared to influence all player evaluations. Since it does not require significantly more time to run, training is faster since calculating evaluations for all players only requires one pass. Player evaluations can just be viewed through an index of the output vector.

### 3.4.2 Feature representation

I split inputs into either Global or Board features. Many of the features are expressed as fractions. This is to keep values bounded which is important as large inputs can prevent the network from converging during training. This also encodes useful information about other players.

- Global Features. These apply to the game as a whole and not a specific position on the board.
    - One-hot encoding of current player turn
    - For each player:
        - Total player armies as a fraction of total armies on the board
        - Total incomes of each player expressed as a fraction of the sum of player incomes currently.
        - Total player territories as a fraction of total territories on the board (42)
        - Current Cards
        - "Defence"
        - For each continent
            - Total player armies in that continent as a fraction of total armies in the continent.
- Board features. These are stored within each territory on the board, containing information on itself and its neighbours.
    - One-hot encoding of territory owner
    - Armies as a fraction of total armies on the board
    - One-hot encoding of continent
    - Is a border between two continents

The "Defence" feature was added after preliminary testing to help the AI learn the concept of defence, which is struggled to do, maybe due to the long-term reward defence gives being harder to identify. It is a handcrafted feature which estimates how defensive the current position is. This is the only instance where I had to build in human knowledge to fix an issue the AI had; however, I was sure to keep the feature global and not to incorporate information about specific threats to the AI.

I use the following algorithm:

1. If the player does not own a continent, return the value 0
2. Else, construct an empty list, $D$.
3. For each owned continent of the player, consider each border territory.
4. For each border territory search all connected owned territories from the continent. Consider each owned territory bordering a territory owned by another player. Add the territory with the least number of armies to $D$, ordered from smallest to largest.

5.  The player defence value, is a weighted arithmetic mean of the armies in the list of size n, divided by the total amount of armies on the board. I use weights of form { n, n-1, … 2, 1}.

The global feature vector has 66 dimensions while each territory has a board feature vector of 14 dimensions.

Before feeding the features into the network, I normalise all except one-hot features through their mean and standard deviation from the training set.

### 3.4.3 Training Data

To train the AI, Risk game data is required for temporal difference learning.

It is desirable to have a large volume of data to expose the AI to a variety of positions and prevent overfitting. This means generating human data is not feasible and I could not find an existing database of human risk games.

The two other options were self-play and AI-generated data. Self-play was not viable due to game-tree search times despite how desirable and promising the concept may to be facilitate autonomous learning. Instead, I used the Angry, Pixie, Cluster, Quo, Killbot and Boscoe AIs inbuilt with Lux to generate 200,000 turn end-states equating to approximately 2000 matches. I chose these specific AIs due to their varying skill and playstyles so a variety of positions would occur that the AI could learn from.

### 3.4.4 TD Learning

Temporal-difference learning is a popular reinforcement learning algorithm which works by training the evaluation function to make it a better predictor of itself at a later timestep by producing an error. This method is desirable because it allows the AI to learn the game itself without human labelling.

I define some variables for the following explanations of temporal difference methods in the context of the project. For a more general explanation, refer to R. Sutton's "Learning to Predict by the Methods of Temporal Differences" [8].

The variable $x_t$ is the game turn end-state at time $t$ represented by global features in a 66-dimensional vector and a graph G defined by each territory and their neighbours on the board, where each node has a 14-dimensional feature vector.

The variable w is a list of parameter vectors that define the weights and biases of the layers in the neural network in the Fully Connected and Graph Convolutional Network layers. The fact it is a list of vectors is an implementation detail and not relevant to the explanation and can easily be thought of as a vector.

$J(x_t, w)_p$ is the evaluation function, the neural network, taking input $x$ and parametrised by $w$. It outputs a 6-dimensional vector of player evaluations. $p$ is for the scalar evaluation for a specific player.

$\nabla J(x_t, w)_p$ is the partial derivative of $J$ with respect to $w$ at $x_t, p$

$r(x_N)_p$ is the terminal reward at the end of the match for each player. It is 0 if the player is dead, and 1 if the player has won.

$d_{t,p} = J(x_{t+1}, w)_p - J(x_t, w)_p$ represents the error between two adjacent evaluations. If a player is dead in the next evaluation, it is overwritten as 0 to represent a loss.

$\alpha$ is the learning rate, $\lambda$ is the credit assignment variable and N is the match (episode) length.

### 3.4.4.1 TD(0)

TD(0) Is one of the simplest temporal difference methods. The general update rule is

$$w = w + \alpha \sum_{t=1}^{N-1} \nabla J(x_t, w) d_t$$

TD(0)

The learning concept is that $w$ will be optimised based on the error between the evaluation of adjacent game states, $d_t$. This is not efficient because the network should be able to infer information about states at least 1-2 full rounds ahead of the first, which equates to 6-12 game-states if t represents 1 player turn and 6 players are playing. In this configuration, it would just predict the evaluation for the very next player turn.

### 3.4.4.2 TD(1)

TD(1) is another simple temporal difference method. The general update rule is

$$w = w + \alpha \sum_{t=1}^{N-1} \nabla J(x_t, w)[r(x_N) - J(x_t, w)]$$

TD(1)

The concept here is that $w$ will be optimised based on the error between the evaluation now and the terminal reward, $r(x_N) - J(x_t, w)$. This is not efficient either because it is impossible to predict too far ahead in the future, for example, trying to predict who will win at the very start of the game. This is due to the complexity and non-determinism in Risk. This results in the AI learning unhelpful features causing erratic evaluations in the early and midgame that confidently predict that one player will win, then switch to confidently predicting another will later.

### 3.4.4.3 TD($\lambda$)

TD($\lambda$) can be viewed as an interpolation between the behaviours of TD(0) and TD(1), which means the issues of both can be mitigated. The general update rule is

$$w = w + \alpha \sum_{t=1}^{N-1} \nabla J(x_t, w) \left[ \sum_{j=t}^{N-1} \lambda^{j-t} d_j \right]$$

TD($\lambda$)

The concept here is that $w$ will be optimised based on an error which assigns more "Blame" to temporal differences, the $d_j$ terms, closer in time to $t$, parametrised by the credit assignment variable $\lambda$ through the exponential decay of $\lambda^{j-t}$. So, the network is optimised to predict the evaluation in the near future, instead of the very next state, or the final state, which is both more useful and realistic than in the other methods. This aids in learning and the extraction of useful features.

### 3.4.4.4 Training

I will use TD($\lambda$) to train the evaluation function to score turn end states for each player. Where each time-step, t represents 1 player turn. Hence the temporal error represents the differences in the evaluation between player turns, and not for actions made within a single turn. This is to aid the AI in looking further ahead into the future and speed up learning.

When using TD learning, initialising the network can help place the evaluation function in a good minimum to speed up training. I have found no difference between a random initialisation, and a bootstrap where I predict the total armies of each player. So, I will use a random initialisation.

To apply the weights to the network, an optimizer is needed. I use the Adadelta algorithm [9]. This is suitable because it reduces learning rate over training aiding convergence while maintaining a separate learning rate for each parameter. This is desirable due to the input data being of varying degrees of significance and with potential learning opportunities for some features being sparse. I am using a learning rate of 0.5 as opposed to the default value 1 because despite normalisation some input values can reach values close to 5 as opposed to the desired range of -1 to 1. This could cause large gradient updates preventing the network from converging.

For training, I collect gradients for each scalar player evaluation, so I implement the following formula.

$$w = w + \alpha \sum_{p=0}^{5} \left( \sum_{t=1}^{N-1} \nabla J(x_t, w)_p \left[ \sum_{j=t}^{N-1} \lambda^{j-t} d_{j,p} \right] \right)$$

Applied TD($\lambda$)

I ran TD($\lambda$) for 3 epochs of the dataset which was lowest value which resulted in confident evaluations and hence limited overfitting. On a GTX 1080, training took approximately 1 hour.

## 3.5 Game Tree Search

The game tree is necessary to consider and evaluate turn end-states in order to return optimal moves. I consider multiple algorithms and due to Risk's very large game tree complexity. I have implemented heuristics to reduce the number of legal moves considered while being careful to only disregard obviously bad moves to prevent hand-coding human knowledge.

It is interesting to note that the AI will search multiple times during a turn due to the need to resync and consider the results of an attack move. To prevent extremely long runtimes, I cut off each search with a parameter, Search Time, and the move chain resulting in the best position is returned. Due to the game-tree complexity of risk, see 2.2, often, only a partial game-tree is constructed.

All the search algorithms generate new game states by applying a subset of the legal moves defined by certain heuristics.

### 3.5.1 Search Heuristics

#### 3.5.1.1 Placing Stage

As shown by M. Wolf's complexity analysis, the placing phase is by far the most complex [6].

The AI only considers placing on territories that border enemies. This is not restrictive since this is what players and AIs do the overwhelming majority of the time, and if the AI wants to have an army behind enemy lines, it can do so in the Fortify Stage.

The AI also only considers $Tp$ territories to place from. This is not too restrictive since players and AIs very rarely place units on more than 3 territories at once. I have found $Tp \geq 2$ to be necessarily so the AI can respond flexibly to attacks.

The final heuristic in the placing phase is dividing the army in groups of $Gp$, then placing down these groups. This means the complexity of the placing stage is not dependant on income. I did not observe this to be significantly restrictive and this hugely reduces the number of placing variations to consider. I have found $Gp = 3$ to be adequate

### 3.5.1.2 Attacking Stage

The AI does not consider attacks where the enemy has a larger number of units. Rarely, a situation does arise where it is desirable to attack where outnumbered and expecting a loss, however I have not observed this to be a problem in testing.

After a successful attack, the AI must decide how many units to move into the newly conquered territory. I reduce the number of possible variations by only considering Ag variations, linearly interpolated from the minimum of 3 and the maximum disregarding duplicates. I found it necessary for $Ga >= 3$ because otherwise the AI did not have the flexibility to execute complicated attack sequences.

### 3.5.1.3 Fortification Stage

Fortifications are an optional move the player could take. This means that all states in the fortification stage are also turn end-states. In combination with the fact that fortifications are generally independent of each other, means I can disconnect the fortification stage from the search, and find which fortifications are desirable by evaluating all fortifications and comparing them to doing nothing, picking the best fortification and repeating until there are no more fortifications which evaluate above doing nothing.

The final heuristic I use is to group fortifications in a way like a successful attack, however I can use more groups because it does not affect the search times and calculating fortifications through the method above is very fast. I found a value of $Gf = 10$ to be effective.

### 3.5.2 Monte Carlo Tree Search

Monte Carlo Tree Search (MCTS) has enjoyed success in recent years for game AI development, particularly with Deep Mind's Alpha Go and Alpha Zero in Go and Chess respectively [2]. MCTS is a greedy and selective search algorithm which allows it to perform better than alternatives in game trees with high branching factors and to implement more expensive evaluation functions such as deep neural networks. Despite how promising this method could be in Risk, I did not use this method due to difficulties batching inputs to the evaluation function which resulted in 10-100x less turn end-states searched than a breath-first method in the same time.

### 3.5.3 Depth First Search

In a depth first search (DFS), the AI would spend most of its time searching sophisticated attack variations which potentially gives the AI an advantage. However, it will regularly miss obvious moves which was apparent in testing.

### 3.5.4 Breath First Search

The breath first search (BFS) solves the problems of a depth-first search. However, it can be short-sighted, only looking a few attacks ahead. This isn't a huge problem though because after every attack the AI searches again, meaning it can still find and execute complicated attack patterns which was observed in testing. I will use a BFS in the AI for final testing and experimentation.

## 3.6 Final Changes

During final testing, I made 2 changes I felt were necessary:

- After the AI has 95% of total armies on the map, the evaluation function changes to be the number of territories owned+1. This is because rarely the AI will not finish off an opponent, keeping it in a continent and preventing it from gaining income, while not observing that it could finish the enemy off due to the depth limited BFS.
- Rarely the AI would place hundreds of units in Australia playing in a totally defensive position, not moving even when an opponent kills all other players and threatens Australia. I suspected this was due the "Defence" feature reaching abnormally high values not observed in the training data, confusing the network . This was fixed by capping the maximum value of the feature before normalisation.

## 3.7 Implementation

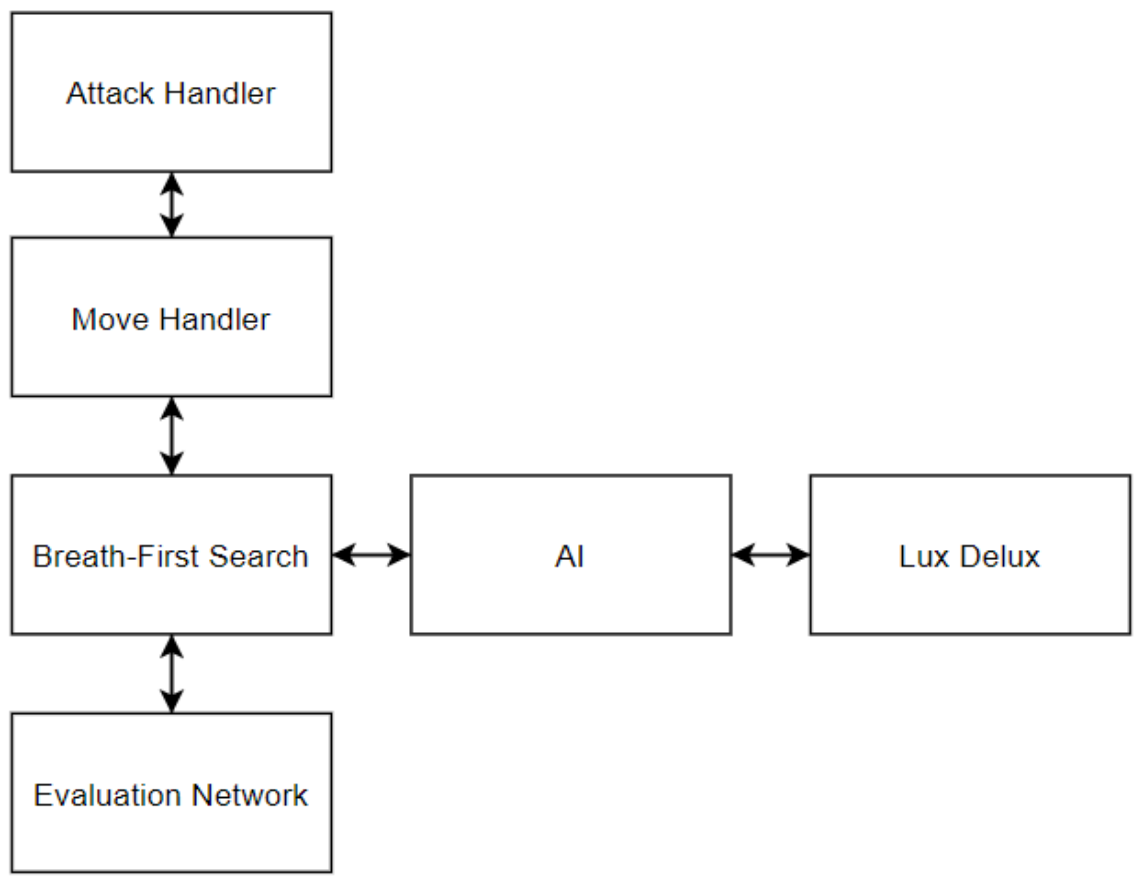

Figure 2: Implementation diagram

I implemented the AI in Python using PyTorch. The AI communicates with Lux Delux through a Java interface which relays all information to the Python process. I use a python object to represent the current state of the game. The move handler generates the subset of legal moves used by the Breath-first search defined by the heuristics previously mentioned. Though I did experiment with multicore utilisation, I decided against it for the final experiments.

### 3.7.1 Parameters

The parameters that control the AI are as follows:

- "Risky": determines how pessimistic the AI is when considering an attack. From experimentation, ranging between 0 and 1, I am using a value of 0.3. Too high results in over-optimistic attacks resulting in lost units and wider branching factors resulting in a lower-depth search. Too low results in very passive behaviour.
- Lambda, $\lambda$: The TD($\lambda$) parameter used in training. Through experimentation, Ranging between 0 and 1, I have found values between 0.7 and 0.95 to have similar results. I will use a value of 0.8.
- Defence Feature Cap: 0.2.
- Neural Network Hyperparameters: These were chosen through experimentation and through the limits of my GPU memory.
    - GCN 1. Out = 60
    - GCN 2. Out = 30
    - FC 1. Out = 60
    - FC 2. Out = 60
    - FC 3. Out = 30
- Search Heuristic Parameters:
    - $Tp = 2$
    - $Gp = 3$
    - $Ga = 3$
    - $Gf = 10$
- Epochs = 3.
- Learning rate $\alpha = 0.5$.
- Search Time = 10s.

## 3.8 Experimental Results

## 3.8.1 Data

To evaluate the skill of the AI at playing Risk, I ran 283 games versus the hardest AIs built into Lux which I could subclass to collect placement statistics. These include Killbot, EvilPixie and Bort, AIs not included in the training data.

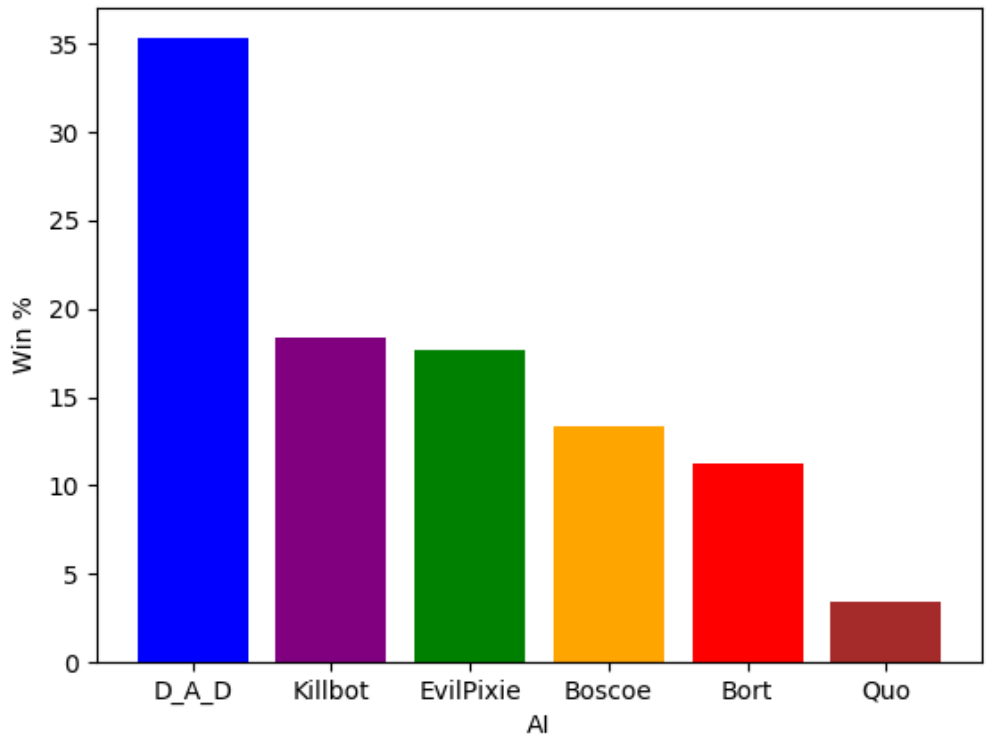

Figure 3: Win % Bar Chart

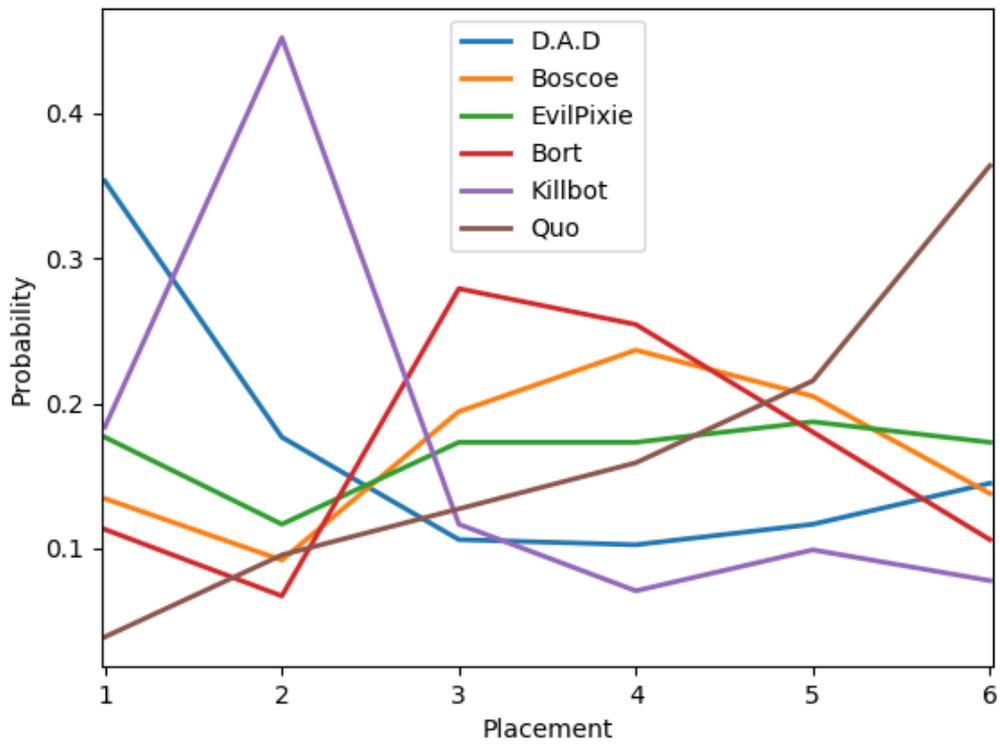

Figure 4: Placement Probability Diagram

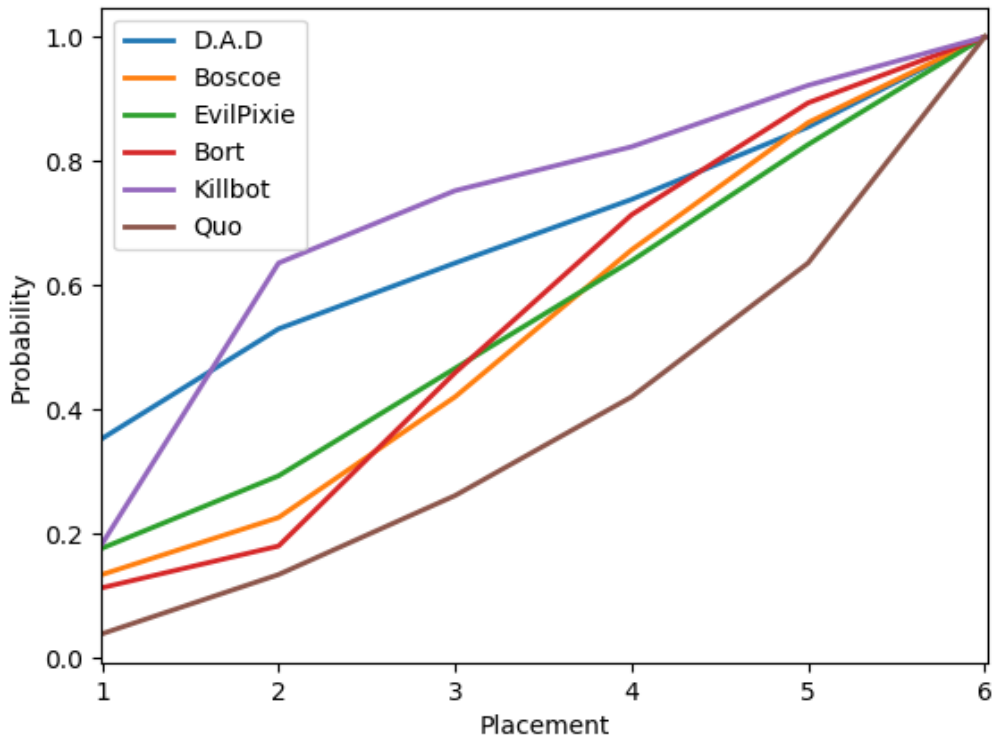

Figure 5: Cumulative Placement Probability Diagram

Figure 3 clearly shows that my AI outperforms the best inbuilt AIs Lux Delux has to offer. It wins 35.3% of the time, nearly double that of Killbot.

### 3.8.2 Behaviour

The AI exhibits "The best defence is a good offense" mentality. This is shown through its overall strategy conquer both Australia and South, then North America while aggressively preventing players, particularly the strongest from holding continents or becoming too powerful. It does this by placing units in or near continents that the AI deems it can take easily. If the AI fails to take South America or Australia, it will act passive, centralising its armies and waiting for an opportunity to take the continents.

Its aggressive playstyle sometimes results in an early loss, but if it can survive after 2 players are eliminated, it is in a good position to win the game. This is shown in Figure 4 where probability for the AI placing 3rd or 4th is lower than placing 5th or 6th.

D.A.D also has a strong endgame, where it attempts to capture territory, prevent opponents from holding continents, while securing and defending continents itself which is what human players do. From observation, it can be as effective as humans in reaching these goals.

This is shown also by Figure 4. If D.A.D has made it such that there is only 1 player left, it can either place 2nd or 1st. The probability of placing 1st in this position is:

$$\frac{p(1^{st})}{p(1^{st}) + p(2^{nd})} = \frac{0.353}{0.353 + 0.185} = 0.656 = 65.5\%$$

This is higher than the 50% chance you would expect if two equally skilled agents played the endgame versus each other, further suggesting that D.A.D has the strongest endgame.

Figure 4 also tends to echo all the playstyles of different AIs. Bort is a similar, slowed down version of Boscoe and hence they have similar placement traces. Quo is the weakest of the hard AIs in lux and this is exploited by the stronger AIs, resulting in a constantly decreasing placement probability.

From the results, Killbot seems to be the strongest of the inbuilt lux AIs. This is shown most clearly in Figure 5, where it places above all other AIs except for $1^{st}$ place. In Figure 4, Killbot has a 45.6% $2^{nd}$ place chance. This indicates its end-game strategy is not the strongest, and this is likely symptomatic of all inbuilt Lux AIs. The set of principles that they follow is not flexible enough to play the endgame, so they have a "Hogwild" state built in. Once past a certain strength threshold, they will do an all-out offensive. However, because the AIs do not search ahead, it may not succeed.

D.A.D is generally insensitive to build-ups of armies on its borders which is on contrast to most AIs and many human players except for some Expert players. The build-up is a natural defensive reaction which helps defend owned continents however it ties up your men which are unable to be used for offensive operations. My AI's offensive playstyle is likely responsible for its insensitivity.

The AI likes to connect its territories together and to centralise its men. These are common ideas shared by other AIs and humans which are nearly always good moves even though they do not result in immediate reward. This is strong evidence that suggests the GCN has learned useful local higher-level features from low-level inputs.

In some positions, the AI can take under a second to perform its entire turn. This is because the game-tree can be explored quickly and fully when the AI does not have many men or is in a locked-in position. However, in many positions the AI will use all the allotted search time. While attacking, this can make some turns take 1-2 minutes to complete.

# 4 Conclusion

From the AIs behaviour and the data, there is clear evidence to support that evaluation function has learned useful high-level features through automatic feature extraction and pattern recognition from low-level input data and uses these features to play at a high level. There is also evidence to suggest that the complexity of the features learned necessitate a deep neural network as opposed to linear regression. Since it has learned entirely from generated AI data and a random initialisation, this method should generalise to any variant of risk on any map that isn't too large. The AI may be even stronger in Risk variants with progressive cards. Since the game moves faster the AI does not need to look as far into the future, helping with both training and searching for moves.

Ultimately, this project has shown it is not only possible to create a Risk playing agent which learns how to play Risk independent of human supervision, but also with minimal handcrafting of its architecture and restriction of its domain as to build in human knowledge of the game. Rather, it learns the overwhelming majority of its knowledge through temporal difference learning from previously played games.

This project also showed that even the simplest Graph Neural Networks can be used to extract useful features on graph data in games.

## 4.1 Areas for further investigation

Due to time constraints I was unable to implement the AI in the most speed and memory efficient way, for example in C++ instead of Python. This would have made the search algorithm significantly faster and allow MCTS to be used, which I feel has potential to enhance the performance of this AI.

As mentioned in 3.1, Creating a range of attack results where the AI would not resync could significantly improve the speed of the AI, resulting in a deeper search tree and a higher playing skill.

I was also not able to gather more detailed statistics to quantify the AIs strategy. Evaluation over time, units/income/territories over time, continents held, average time territory and individual army sizes would be interesting statistics to have and would allow for more in-depth analysis to facilitate improvements to methods used in this project.

# Acknowledgements

This project was inspired by the recent success of AlphaGo and the AI is named after my Dad, an expert Risk player who helped me with bouncing ideas for the project.